\documentclass[11pt]{article}

\usepackage[margin=1in]{geometry}
\usepackage{amsmath,amssymb}
\usepackage{graphicx}
\usepackage{booktabs}
\usepackage{array}
\usepackage{multirow}
\usepackage[hidelinks]{hyperref}
\usepackage{authblk}
\usepackage{caption}
\usepackage{enumitem}

\title{FlyHash for LLM Agent Memory: An Empirical Validation and a New Application}
\author[1]{Mohan Bheemadolu\thanks{Code: \url{https://github.com/mohanji/flyhash-agent-memory}\\ Archive (DOI): \url{https://doi.org/10.5281/zenodo.23903302}}}
\author[2]{Chrisogonas O. Odhiambo}
\affil[1]{Independent Researcher. Email: \texttt{mohan.bmdl@gmail.com}}
\affil[2]{Email: \texttt{chrisogonas@gmail.com}}
\date{}

\begin{document}
\maketitle

\begin{abstract}
Large language model (LLM) agents increasingly rely on external memory
stores to retain information across conversations, typically via dense
vector embeddings searched with approximate nearest-neighbor methods. The
fruit fly's olfactory circuit computes a locality-sensitive hash of sensory
input using a strikingly different mechanism --- sparse random projection
followed by winner-take-all competition --- and prior work (Dasgupta et al.,
2017) showed this scheme, dubbed FlyHash, preserves neighbor structure
better per bit than conventional random-hyperplane hashing (SimHash) on
several benchmarks. We apply FlyHash, to our knowledge for the first
time, to LLM agent memory retrieval, and ask three questions a
practitioner actually needs answered: does the quality advantage hold
under a correctly matched-bit protocol (it does, robustly, across 10
seeds and four datasets spanning two modalities, with one specific,
previously undocumented comparison pitfall identified and corrected
along the way); does that advantage translate into real memory and
latency savings, including against the dense-vector indexing (HNSW)
practitioners default to today (no: at equal bits FlyHash's codes are
larger and slower to compare than SimHash's, and HNSW dominates both hash
methods at ordinary scale; a sparse-set comparison we introduce recovers
most but not all of a naive latency gap, leaving hashing's practical case
narrower and more specific than the quality result alone implies); and does the advantage survive genuine, non-circular
agent-memory relevance labels rather than the nearest-neighbor-by-
embedding proxy this literature otherwise relies on (it does). We report
all three answers precisely, including the two that narrow rather than
flatter the headline result, and release our implementation, a working
agent-memory demo, and a full account of the methodological pitfalls we
found and corrected.
\end{abstract}

\section{Introduction}

Large language model agents that operate over long or repeated
interactions increasingly need memory: a way to store facts, preferences,
or prior context and retrieve the relevant pieces on demand
(Chhikara et al., 2025). The dominant
approach embeds each memory as a dense vector and retrieves by approximate
nearest-neighbor search over an index such as FAISS (Johnson et al., 2021; see also the RAG survey of
Zhao et al., 2024),
though hybrid retrieval architectures combining dense and lexical signals
are an active and growing alternative. This works well but
scales in memory and compute with the size and dimensionality of the
stored embeddings, motivating interest in more compact representations ---
in particular, hashing schemes that compress an embedding into a short
binary code while preserving enough neighbor structure to remain useful
for retrieval.

One such scheme, introduced by Dasgupta, Stevens, and Navlakha (2017) and
directly inspired by neurobiology, computes a locality-sensitive hash the
way the fruit fly's olfactory circuit is believed to encode odors: roughly
50 receptor-neuron activations are randomly and sparsely projected into
approximately 2{,}000 ``Kenyon cells,'' of which only the top $\sim$5\% most
active are allowed to ``fire,'' producing a sparse binary code. The original
paper showed that this scheme --- which we refer to as FlyHash, following
subsequent literature --- preserves nearest-neighbor structure better, at a
matched number of output bits, than SimHash (Charikar, 2002), the standard
random-hyperplane hashing baseline, on several benchmark datasets including
SIFT, GloVe, and MNIST.

Despite this result and several follow-up papers extending the algorithm
(surveyed in Section~\ref{sec:related}), we are not aware of any prior work
applying FlyHash-style hashing specifically to LLM agent memory retrieval
--- a setting that is, in several respects, a natural fit: agent memories
are frequently decomposed into short, numerous, semantically embedded
units retrieved by similarity (Hu et al., 2026; Du, 2026; Zeng et al.,
2024) --- the regime this paper's evaluation targets, though we note that
recent work increasingly favors structured or hierarchical retrieval over
flat similarity search for longer-horizon tasks (Cao et al., 2026; Hu et
al., 2026).

This paper is organized around three research questions, each answered
precisely rather than in the aggregate, because --- as we show below ---
the answers are not all the same shape:

\begin{itemize}[leftmargin=*]
\item \textbf{RQ1.} Under a protocol that correctly holds FlyHash's and
SimHash's \emph{active} bit counts equal (Section~\ref{sec:protocol}),
does FlyHash's reported per-bit retrieval-quality advantage over SimHash
hold for data representative of LLM agent memory, robustly across seeds
and datasets? (Sections~\ref{sec:main-result}--\ref{sec:hyperparam};
answer: yes, with one specific, previously undocumented comparison
pitfall that must be avoided to get there --- Appendix~\ref{sec:appendixA}.)
\item \textbf{RQ2.} Does that quality advantage translate into an actual
resource advantage --- lower memory, lower query latency --- in a
realistic deployment, including against the dense-vector indexing
methods (HNSW) practitioners use by default today?
(Sections~\ref{sec:memlatency}--\ref{sec:sparse}, \ref{sec:hnsw-systems};
answer: not by default, and not relative to HNSW at ordinary scale ---
hashing's resource case is narrower than the quality result alone would
suggest.)
\item \textbf{RQ3.} Does the quality advantage hold under genuine,
non-circular agent-memory relevance labels, rather than the
nearest-neighbor-by-embedding proxy used everywhere else in this
literature (including the original FlyHash paper)?
(Section~\ref{sec:realistic-benchmark}; answer: yes.)
\end{itemize}

In service of these questions, the paper's contributions are: (1) the
first application, to our knowledge, of FlyHash to LLM agent memory
retrieval, with a working end-to-end system
(Sections~\ref{sec:agent-method}, \ref{sec:agent-demo}); (2) the
matched-active-bit protocol itself, stated precisely enough to prevent a
specific conflation we discovered by reproducing a discrepancy with a
published result (Section~\ref{sec:protocol}, Appendix~\ref{sec:appendixA});
(3) statistical and cross-dataset validation beyond the original paper
--- 10 seeds, four datasets across two modalities, a 36-configuration
hyperparameter sweep, and a paired statistical analysis
(Sections~\ref{sec:main-result}--\ref{sec:hyperparam}); and (4) an
honest, multi-angle resource and deployment analysis that does not stop
at the quality result --- brute-force memory/latency costs, a sparse-set
comparison, a direct HNSW systems baseline, and a relevance-label
upgrade (Sections~\ref{sec:memlatency}--\ref{sec:sparse},
\ref{sec:hnsw-systems}--\ref{sec:realistic-benchmark}).

The remainder of this paper is organized as follows.
Section~\ref{sec:related} situates this work relative to FlyHash's
original paper and its algorithmic follow-ups. Section~\ref{sec:method}
describes the method, evaluation protocol, datasets, and statistical
procedure. Section~\ref{sec:results} presents results.
Section~\ref{sec:discussion} discusses their implications.
Section~\ref{sec:limitations} states limitations and future work.
Appendix~\ref{sec:appendixA} gives a full account of the reproducibility
investigation that motivated the corrected protocol in
Section~\ref{sec:protocol}.

\section{Related Work}
\label{sec:related}

\textbf{The original algorithm.} Dasgupta, Stevens, and Navlakha (2017)
introduced the algorithm this paper builds on, showing that a
neurally-plausible circuit --- sparse random projection into a much
higher-dimensional space, followed by winner-take-all sparsification ---
functions as a locality-sensitive hash competitive with, and often
superior to, dense random-hyperplane hashing (SimHash; Charikar, 2002) at
a matched number of output bits. Their evaluation covered SIFT
descriptors, GloVe word vectors, and MNIST images. This paper directly
reuses their algorithm and their choice of evaluation datasets (GloVe and
MNIST; we do not evaluate on SIFT --- see Section~\ref{sec:limitations}),
applying it to a problem domain they did not consider.

\textbf{DenseFly.} Sharma and Navlakha (2018) extended the base algorithm
with ``DenseFly,'' which densifies FlyHash's sparse output as a
post-processing step to further improve similarity search accuracy. This
is a genuine algorithmic improvement over plain FlyHash that we do not
implement here; we discuss it as natural future work in
Section~\ref{sec:limitations}.

\textbf{BioHash.} Ryali, Hopfield, Grinberg, and Krotov (2020) replaced
FlyHash's \emph{random} sparse projection with a \emph{learned} one,
trained with a biologically-plausible local learning rule, and reported
improved retrieval accuracy over the random-projection version across
several datasets. Their paper also established mean Average Precision
(mAP) as the standard evaluation metric for this line of work, which we
adopt in Section~\ref{sec:protocol}. We note for transparency that
BioHash's learned projection is reported to outperform the plain FlyHash
evaluated in this paper; our contribution is not a better hash function,
but an application and validation of the simpler, random-projection
version, which requires no training data or learning procedure --- a
practically relevant trade-off for a lightweight agent-memory component.

\textbf{FlyNN.} Most recently, work applying the fly olfactory circuit to
federated nearest-neighbor classification (``FlyNN,'' AAAI 2022)
illustrates the breadth of problems this biological mechanism has been
adapted to beyond pure hashing, extending it to a federated learning
setting.

\textbf{Existing implementations.} A general-purpose \texttt{FlyHash}
package is available on PyPI (Huang, MIT license), providing a
NumPy-compatible implementation of the base algorithm for
small-to-medium input dimensionalities. Its existence confirms that an
implementation of the algorithm itself is not a novel contribution; our
implementation exists to support the specific experiments in this paper
rather than to serve as a general-purpose library.

\textbf{Graph-based approximate nearest-neighbor search.} A different
line of work achieves fast approximate retrieval not by compressing each
vector into a short code, but by building a navigable graph over the
full-precision (or lightly quantized) vectors and searching it via
greedy graph traversal. HNSW (Hierarchical Navigable Small World graphs;
Malkov and Yashunin, 2016/2020) is the dominant practical method in this
family and underlies most production vector-search systems in use today
(e.g., FAISS's \texttt{IndexHNSW}, and the default indexes in several
vector databases including Milvus, Pinecone, Weaviate, and pgvector;
Idrees et al., 2026). HNSW and hashing schemes like FlyHash and SimHash
target different points in the retrieval design space: HNSW optimizes
for sublinear query time and high recall at the cost of storing richer
per-item structure (full or near-full vectors plus graph edges), while
FlyHash and SimHash optimize for extreme per-record compactness --- a
short binary code compared via cheap Hamming distance --- at the cost of
the brute-force, linear-scan comparison this paper evaluates throughout.
Our main results (Sections~\ref{sec:results}.1--\ref{sec:results}.4)
isolate hashing-function quality by comparing FlyHash against SimHash at
matched bit budgets, rather than against HNSW, since the two are
different design points (compact-code quality versus indexed graph
search); we separately report a direct, explicitly-labeled systems
comparison against HNSW in Section~\ref{sec:results} for readers asking
the practical ``which would I actually deploy'' question.

\textbf{LLM memory architectures and retrieval-augmented generation.} A
large and rapidly growing body of work designs memory systems for LLM
agents directly, distinct from the hashing-algorithm literature surveyed
above: production-oriented long-term memory frameworks (Chhikara et al.,
2025), structural memory representations spanning chunks, triples, and
summaries (Zeng et al., 2024), and recent surveys of agent memory
mechanisms and evaluation (Du, 2026). This literature's central recent
finding --- that flat, independent similarity retrieval over text
snippets can suffer from redundancy and relational blindness at longer
horizons, motivating structured or hierarchical retrieval instead (Hu et
al., 2026; Cao et al., 2026) --- is complementary to this paper's
contribution: we evaluate hashing methods within the flat similarity-based
retrieval regime that remains a widely used baseline (e.g., Mem0-style
systems), rather than proposing or evaluating a structured memory
architecture; combining FlyHash-style hashing with structured retrieval is
natural future work (Section~\ref{sec:limitations}).

\textbf{Vector quantization and alternative hashing families.} A separate
line of work compresses dense vectors via quantization rather than binary
hashing. Product Quantization (Jégou, Douze, \& Schmid, 2010) and its
optimized variants (Ge et al., 2013) achieve higher retrieval accuracy
than binary hashing at comparable memory budgets by preserving continuous
distance estimates via asymmetric lookup tables rather than collapsing to
Hamming codes; recent methods such as RaBitQ extend this advantage to
extreme, 1-bit-per-dimension compression (Gao \& Long, 2024). We do not
evaluate against this family here, since our focus is a head-to-head
comparison between two hashing paradigms under a matched-bit protocol, but
it represents a stronger baseline than either hash method at most
operating points --- consistent with the HNSW result in
Section~\ref{sec:hnsw-systems}. Within the hashing family itself, more
sophisticated locality-sensitive schemes such as Cross-Polytope LSH
achieve a better asymptotic query exponent than random-hyperplane hashing
(Laarhoven, 2020), but this advantage only materializes at large scale and
high ambient dimension; at the moderate dataset sizes evaluated here
(Table~\ref{tab:datasets}), hyperplane-based SimHash remains competitive
or superior in wall-clock performance due to lower per-hash computational
overhead (Cai, 2016; Laarhoven, 2020), motivating our choice of standard
SimHash, rather than Cross-Polytope LSH, as the baseline.

\textbf{Sparse lexical retrieval.} A different line of ``sparse retrieval''
work --- BM25, and learned sparse lexical models such as SPLADE (Formal,
Lassance, Piwowarski, \& Clinchant, 2024) --- operates directly on text via
sparse term-weight vectors over a vocabulary, which is a distinct notion
of sparsity from the sparse \emph{binary codes} FlyHash derives from dense
embeddings via winner-take-all competition. These lexical methods are a
complementary, not competing, approach, well-suited to exact keyword
matching rather than semantic similarity over learned embeddings, and are
outside this paper's scope of embedding-based hashing methods.

\textbf{This paper's position.} None of the above apply FlyHash-style
hashing to LLM agent memory retrieval, and none report the depth of
statistical validation (multi-seed variance, multi-dataset,
multi-modality, hyperparameter ablation) presented in
Section~\ref{sec:results}. The gap this paper fills is at the
intersection of a new, currently relevant application and a level of
empirical rigor --- including a documented and corrected methodological
pitfall --- that the base algorithm's literature does not yet provide.

\section{Method}
\label{sec:method}

\subsection{FlyHash and SimHash}

Both hashing schemes take a dense input vector $x \in \mathbb{R}^d$ and
produce a binary output code.

\textbf{SimHash} (Charikar, 2002) draws a dense random matrix
$W \in \mathbb{R}^{k \times d}$ with entries sampled i.i.d.\ from a
standard normal distribution, and outputs
\[
h_{\text{sim}}(x) = \operatorname{sign}(Wx) \in \{0,1\}^k.
\]
Every one of the $k$ output bits is thresholded independently; the code
is dense in the sense that each bit is set to 1 with probability
approximately $1/2$.

\textbf{FlyHash} (Dasgupta et al., 2017) instead draws a \emph{sparse}
random projection matrix $P \in \mathbb{R}^{m \times d}$, where each row
(each ``Kenyon cell'') is connected to only a random subset of
$\lceil s \cdot d \rceil$ input dimensions ($s$ is the \emph{sparsity},
fixed at 0.10 throughout this paper, following the fly's own
connectivity, where each Kenyon cell integrates inputs from roughly
5--12\% of input channels; Caron, Ruta, Abbott, \& Axel, 2013), with
weights drawn uniformly from $\{-1,+1\}$ on those
connections and zero elsewhere. The projection $Px$ is then passed
through a $k$-winners-take-all operation: only the $k$ largest of the $m$
projected values are set to 1, all others to 0:
\[
h_{\text{fly}}(x) = \operatorname{WTA}_k(Px) \in \{0,1\}^m, \qquad
\lVert h_{\text{fly}}(x) \rVert_0 = k.
\]
We refer to $m$ as the \textbf{pool size} (the total number of Kenyon
cells, fixed at 2{,}000 in our experiments unless otherwise stated) and
$k$ as the \textbf{bit budget} (the number of winning, active bits). This
notational distinction --- $k$ versus $m$ --- is deliberately explicit,
because conflating the two is the source of the reproducibility
discrepancy described in Section~\ref{sec:protocol} and
Appendix~\ref{sec:appendixA}.

\subsection{Evaluation protocol}
\label{sec:protocol}

\textbf{Metric.} Following BioHash (Ryali et al., 2020), we evaluate
retrieval quality with mean Average Precision at $R$ (mAP@$R$). For a
query $q$, let the \emph{relevant set} be its $R$ true nearest neighbors
under exact Euclidean distance on the original (unhashed) embeddings.
Average Precision for $q$ is
\[
\text{AP}(q) = \frac{1}{R}\sum_{\ell=1}^{R} \text{Precision}(\ell) \cdot
\text{Rel}(\ell),
\]
where $\text{Rel}(\ell) = 1$ if the item ranked $\ell$-th by Hamming
distance under the hash code is in the relevant set, and
$\text{Precision}(\ell)$ is the fraction of relevant items among the top
$\ell$ ranked results. mAP@$R$ is the mean of $\text{AP}(q)$ over all
queries. Unlike plain recall@$k$, mAP@$R$ rewards ranking true neighbors
earlier, not merely including them somewhere in the top $R$.

\textbf{Matched-active-bit (quality-per-active-bit) protocol.} The
central methodological point of this paper concerns what it means to
compare FlyHash and SimHash ``at the same number of bits.'' There are at
least three distinct, non-interchangeable notions of parity one could
hold fixed:

\begin{enumerate}
\item \textbf{Equal active/output bits} --- the number of bits that are
actually compared by Hamming distance per query ($k$ for both methods).
This is the notion we use throughout
Sections~\ref{sec:main-result}--\ref{sec:hyperparam}, and we call it the
\textbf{matched-active-bit} protocol to avoid the vaguer, overloaded term
``fair comparison.''
\item \textbf{Equal physical storage} --- the number of bytes each
method's code occupies at rest, which depends on \emph{representation}
(dense vs.\ packed vs.\ sparse-index), not just $k$
(Sections~\ref{sec:memlatency}--\ref{sec:sparse}).
\item \textbf{Equal query-time/resource budget} --- the number of bytes
moved or compared per query, which can differ from storage when a
method's on-disk and in-memory representations differ
(Sections~\ref{sec:memlatency}--\ref{sec:sparse}, \ref{sec:hnsw-systems}).
\end{enumerate}

Conflating these is the specific pitfall this paper is built around.
Concretely, under the matched-active-bit protocol, for each bit budget
$k$ we evaluate FlyHash with pool size $m$ fixed at a large value
(2{,}000, unless stated otherwise) and the winner-take-all fraction set
to $k/m$, and we evaluate SimHash with exactly $k$ output bits --- so
both methods produce the same number of \emph{active} bits, isolating
the quality of the hash \emph{function} from either method's absolute
output size. This is not the naive way to run the comparison: a naive
implementation might instead fix FlyHash's \emph{total} output width
(i.e., set $m$ to whatever ``bits'' value is being swept) and hold the
winner-take-all \emph{percentage} fixed (e.g., 5\%, matching the fly's
own biology) --- in which case FlyHash's \emph{effective} number of
active bits at a given nominal ``bits'' value is only a small fraction
of that value, while SimHash is given the full nominal count as
genuinely active bits. This conflation of $k$ (notion 1) with $m$
silently and substantially disadvantages FlyHash, and is distinct from
the storage/resource conflations addressed later (notions 2 and 3,
Sections~\ref{sec:memlatency}--\ref{sec:sparse}). We discovered the
first pitfall by reproducing a discrepancy with a published result; the
full diagnostic process, including hypotheses we initially pursued and
later ruled out, is given in Appendix~\ref{sec:appendixA}.

\subsection{Datasets}

We evaluate on four independent real datasets spanning two modalities,
summarized in Table~\ref{tab:datasets}.

\begin{table}[h]
\centering
\caption{Datasets used for the main comparison (Section~\ref{sec:main-result}).}
\label{tab:datasets}
\small
\begin{tabular}{@{}p{2.6cm}p{1.7cm}p{1.4cm}p{1.1cm}p{6.2cm}@{}}
\toprule
Dataset & Modality & Records & Dim. & Role \\
\midrule
GloVe (glove-wiki-gigaword-100) \cite{pennington2014glove} & Text (word vectors) & 5{,}000 & 100 &
Primary dataset; discovery of the fair-comparison pitfall (Sec.~\ref{sec:protocol}, App.~\ref{sec:appendixA}); statistical (Sec.~\ref{sec:statprotocol}) and hyperparameter (Sec.~\ref{sec:hyperparam}) validation \\[6pt]
MiniLM (all-MiniLM-L6-v2) \cite{reimers2019minilm} & Text (sentence embeddings) & 100 & 384 &
Validates the exact embedding type used in the agent memory demonstration (Sec.~\ref{sec:agent-method}) \\[6pt]
scikit-learn \texttt{digits} \cite{pedregosa2011sklearn} & Image & 1{,}797 & 64 (8$\times$8) &
Genuine, if small, image dataset used where full MNIST was not reachable during development \\[6pt]
MNIST (via OpenML) \cite{lecun1998mnist} & Image & 5{,}000 (of 70{,}000) & 784 (28$\times$28) &
Full-scale image benchmark used in the original FlyHash paper \\
\bottomrule
\end{tabular}
\end{table}

All embeddings are L2-normalized prior to hashing where applicable
(GloVe, MiniLM). MNIST and \texttt{digits} pixel intensities are scaled
to $[0,1]$.

\textbf{Data availability.} All four datasets are public and loaded
programmatically, and none of them is redistributed in this paper's code
release: GloVe via \texttt{gensim}'s model downloader
(\url{https://nlp.stanford.edu/projects/glove/}); MiniLM sentence
embeddings via the \texttt{sentence-transformers} library, which
downloads the pretrained checkpoint from
\url{https://huggingface.co/sentence-transformers/all-MiniLM-L6-v2};
scikit-learn's bundled \texttt{digits} dataset via
\texttt{sklearn.datasets.load\_digits()}; and MNIST via OpenML
(\url{https://www.openml.org/d/554}), fetched with
\texttt{sklearn.datasets.fetch\_openml()}. Exact loading code is in the
repository linked on the title page. The labeled agent-memory benchmark of
Section~\ref{sec:realistic-benchmark} (6{,}000 memories, 150 queries, relevance
labels) is released as JSONL files in the same repository
(\texttt{datasets/agent\_memory\_v1/}), together with an offline demo, tests,
and a \texttt{REPRODUCING.md} guide mapping each table to a command. The
release is archived on Zenodo (release v1.0; DOI
\href{https://doi.org/10.5281/zenodo.23903302}{10.5281/zenodo.23903302}, which
resolves to the latest version).

\subsection{Statistical protocol}
\label{sec:statprotocol}

Both FlyHash's sparse projection matrix and SimHash's dense hyperplane
matrix are randomly initialized, so a single evaluation reflects one
particular random draw, not the expected behavior of the method. For our
main results (Section~\ref{sec:main-result}) and hyperparameter ablation
(Section~\ref{sec:hyperparam}), we repeat every evaluation across $N$
independent random seeds (10 for the main results, 3 for the larger
hyperparameter sweep, given its greater number of configurations),
varying \emph{only} the projection matrices' initialization across seeds
while holding the dataset and the sampled set of queries fixed. This
isolates projection-matrix randomness as the source of measured
variance, rather than conflating it with variance from query sampling.
We report the mean and standard deviation of mAP@$R$ across seeds, and,
as a simple and intentionally conservative robustness check, whether the
gap between FlyHash's and SimHash's mean mAP exceeds twice their
combined standard deviation
($\Delta > 2\sqrt{\sigma_{\text{fly}}^2 + \sigma_{\text{sim}}^2}$). We
present this as a sanity check against attributing an observed gap to
projection-matrix noise, not as a substitute for a formal statistical
significance test.

Because FlyHash and SimHash are evaluated on the \emph{same} 10 seeds
(same sampled queries, same train/query split, differing only in which
method's projection matrix that seed initializes), the two methods' mAP
values at a given bit budget are naturally paired, not independent
samples --- each seed contributes one paired difference
$d_i = \text{mAP}_{\text{fly},i} - \text{mAP}_{\text{sim},i}$. For our
primary GloVe result (the first block of Table~\ref{tab:main-results},
and Table~\ref{tab:paired-stats} below) we additionally report
a paired analysis: the mean paired difference with both a $t$-interval
and a bootstrap (10,000 resamples) 95\% confidence interval, the paired
effect size (Cohen's $d_z$), and two nonparametric tests that do not
assume normality --- the Wilcoxon signed-rank test and a sign-flip
permutation test (10,000 permutations). This is a strictly stronger
statement than the ``robust'' heuristic above, which compares marginal
variances and discards the pairing.

\subsection{Agent memory integration}
\label{sec:agent-method}

To validate the practical use case motivating this work, we implement a
minimal LLM agent memory system: a plain text \texttt{remember(text)} /
\texttt{recall(query, k)} interface wraps a sentence-transformer
embedding model and a FlyHash-backed store, so that an agent interacts
only with text, not embeddings or hash codes directly. We wire this into
a live tool-use loop with a general-purpose LLM exposing two tools ---
\texttt{save\_memory}, called when the model judges the user has shared a
fact worth remembering, and \texttt{recall\_memory}, called when
answering a question that may depend on a previously shared fact. In a
representative interaction, a user states two facts in one turn (``My
favorite language is Python and I'm allergic to peanuts''), asks an
unrelated question in the next turn, and then asks ``What should I avoid
eating?'' in a third turn; the agent correctly answers ``peanuts'' by
calling \texttt{recall\_memory}, retrieving the relevant fact via
FlyHash-based Hamming-distance ranking over stored memory embeddings from
several turns earlier. This demonstrates the full pipeline functioning
end-to-end, independent of the benchmark-style evaluation in
Section~\ref{sec:results}.

\section{Results}
\label{sec:results}

\subsection{Main result: corrected comparison across four datasets}
\label{sec:main-result}

Table~\ref{tab:main-results} reports mAP@$R$ (mean $\pm$ standard
deviation over 10 random seeds) for FlyHash and SimHash at five bit
budgets, on each of the four datasets described in Section~\ref{sec:method},
under the corrected protocol of Section~\ref{sec:protocol}. ``Robust''
indicates whether the gap between the two methods' means exceeds twice
their combined standard deviation (Section~\ref{sec:statprotocol}).

\begin{table}[h]
\centering
\caption{Main results: mAP (mean $\pm$ std over 10 seeds) for FlyHash vs.\ SimHash across four datasets and five bit budgets.}
\label{tab:main-results}
\small
\begin{tabular}{crlll}
\toprule
Dataset & $k$ (bits) & FlyHash & SimHash & Robust \\
\midrule
\multirow{5}{*}{GloVe, mAP@100}
 & 16  & \textbf{0.196 $\pm$ 0.007} & 0.047 $\pm$ 0.006 & Yes \\
 & 32  & \textbf{0.314 $\pm$ 0.009} & 0.096 $\pm$ 0.009 & Yes \\
 & 64  & \textbf{0.449 $\pm$ 0.008} & 0.185 $\pm$ 0.008 & Yes \\
 & 128 & \textbf{0.585 $\pm$ 0.006} & 0.316 $\pm$ 0.006 & Yes \\
 & 256 & \textbf{0.705 $\pm$ 0.006} & 0.469 $\pm$ 0.007 & Yes \\
\addlinespace
\multirow{5}{*}{MiniLM, mAP@20}
 & 16  & \textbf{0.310 $\pm$ 0.013} & 0.214 $\pm$ 0.026 & Yes \\
 & 32  & \textbf{0.408 $\pm$ 0.018} & 0.266 $\pm$ 0.022 & Yes \\
 & 64  & \textbf{0.537 $\pm$ 0.030} & 0.355 $\pm$ 0.024 & Yes \\
 & 128 & \textbf{0.656 $\pm$ 0.016} & 0.466 $\pm$ 0.027 & Yes \\
 & 256 & \textbf{0.753 $\pm$ 0.013} & 0.587 $\pm$ 0.016 & Yes \\
\addlinespace
\multirow{5}{*}{\texttt{digits}, mAP@50}
 & 16  & \textbf{0.456 $\pm$ 0.015} & 0.119 $\pm$ 0.031 & Yes \\
 & 32  & \textbf{0.593 $\pm$ 0.010} & 0.236 $\pm$ 0.039 & Yes \\
 & 64  & \textbf{0.699 $\pm$ 0.011} & 0.397 $\pm$ 0.023 & Yes \\
 & 128 & \textbf{0.748 $\pm$ 0.005} & 0.535 $\pm$ 0.021 & Yes \\
 & 256 & \textbf{0.806 $\pm$ 0.007} & 0.664 $\pm$ 0.009 & Yes \\
\addlinespace
\multirow{5}{*}{MNIST, mAP@100}
 & 16  & \textbf{0.276 $\pm$ 0.009} & 0.064 $\pm$ 0.014 & Yes \\
 & 32  & \textbf{0.382 $\pm$ 0.010} & 0.133 $\pm$ 0.012 & Yes \\
 & 64  & \textbf{0.469 $\pm$ 0.011} & 0.237 $\pm$ 0.014 & Yes \\
 & 128 & \textbf{0.527 $\pm$ 0.008} & 0.351 $\pm$ 0.009 & Yes \\
 & 256 & \textbf{0.563 $\pm$ 0.004} & 0.450 $\pm$ 0.010 & Yes \\
\bottomrule
\end{tabular}
\end{table}

\begin{table}[h]
\centering
\caption{Paired statistical analysis, GloVe: mean FlyHash-minus-SimHash mAP difference per seed (10 seeds).}
\label{tab:paired-stats}
\small
\begin{tabular}{rlllrrr}
\toprule
$k$ (bits) & Mean diff & 95\% CI ($t$) & 95\% CI (bootstrap) & Cohen's $d_z$ & Wilcoxon $p$ & Perm.\ $p$ \\
\midrule
16  & 0.1485 & [0.1421, 0.1549] & [0.1437, 0.1539] & 16.6 & 1.95e-03 & 2.30e-03 \\
32  & 0.2179 & [0.2073, 0.2284] & [0.2095, 0.2267] & 14.8 & 1.95e-03 & 2.30e-03 \\
64  & 0.2642 & [0.2550, 0.2734] & [0.2568, 0.2719] & 20.6 & 1.95e-03 & 2.30e-03 \\
128 & 0.2691 & [0.2643, 0.2738] & [0.2654, 0.2732] & 40.8 & 1.95e-03 & 2.30e-03 \\
256 & 0.2354 & [0.2284, 0.2424] & [0.2299, 0.2414] & 24.1 & 1.95e-03 & 2.30e-03 \\
\bottomrule
\end{tabular}
\end{table}

$t$-interval and bootstrap CIs agree closely at every bit budget, and
Cohen's $d_z$ is an order of magnitude past the conventional ``large''
threshold of 0.8 throughout. The Wilcoxon and permutation $p$-values sit
at $1/2^{10} \approx 0.00195$, the smallest value either test can report
with $n=10$ paired seeds --- FlyHash wins on all 10 seeds at every
budget, so this is the resolution floor of the test, not a computational
artifact. This paired analysis, not the ``gap $>$ 2$\times$combined std''
heuristic alone, is what we consider the primary evidence of a robust
effect; the heuristic is retained in Table~\ref{tab:main-results} as a
quick, consistent summary across all four datasets.

\begin{figure}[h]
\centering
\includegraphics[width=0.75\textwidth]{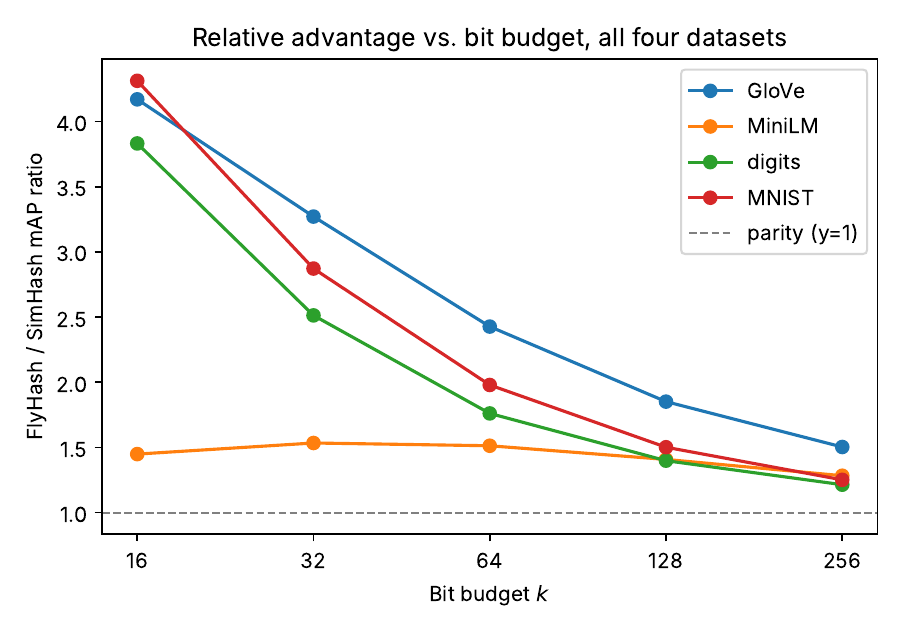}
\caption{FlyHash/SimHash mAP ratio vs.\ bit budget, all four datasets.
Every dataset's curve is monotonically decreasing and stays above the
parity line ($y=1$) across the full range tested, with GloVe and MNIST
showing the steepest initial advantage and MiniLM the flattest curve (a
smaller relative gap that still never crosses parity).}
\label{fig:relative-advantage}
\end{figure}

FlyHash outperforms SimHash at every bit budget, on every dataset, with
every gap classified robust --- 20 independent (dataset, bit-budget)
comparisons, zero exceptions. A consistent secondary pattern is also
visible across all four datasets: the \emph{relative} advantage
($\text{mAP}_{\text{fly}} / \text{mAP}_{\text{sim}}$) is largest at the
smallest bit budget and shrinks as $k$ grows --- for example, on GloVe,
FlyHash is $4.2\times$ SimHash at $k=16$ but only $1.5\times$ at $k=256$.
Figure~\ref{fig:relative-advantage} plots this ratio against bit budget
for all four datasets at once, making the shared pattern visible across
modalities. We return to the significance of this pattern in
Section~\ref{sec:discussion}.

\subsection{Hyperparameter ablation}
\label{sec:hyperparam}

FlyHash has two hyperparameters left fixed throughout
Section~\ref{sec:main-result}: sparsity $s$ (fraction of input dimensions
each Kenyon cell samples from; 0.10 by default, following the fly's own
connectivity) and pool size $m$ (2{,}000 by default). To check whether
the Section~\ref{sec:main-result} result depends on these specific
choices, we sweep $s \in \{0.05, 0.10, 0.20\}$ and
$m \in \{500, 1000, 2000, 4000\}$ --- 12 combinations --- at three
representative bit budgets ($k \in \{16, 64, 256\}$), on GloVe, with 3
seeds per configuration (fewer than Section~\ref{sec:main-result}'s 10,
given the larger number of configurations swept here).

\begin{table}[h]
\centering
\caption{FlyHash mAP@100 (mean $\pm$ std over 3 seeds) across the
sparsity $\times$ pool-size grid, at three bit budgets. SimHash's mAP@100
at the same $k$ is given as the baseline for each sub-table.}
\label{tab:hyperparam}
\small
\begin{tabular}{ccccc}
\toprule
\multicolumn{5}{c}{$k=16$ (SimHash baseline: $0.046 \pm 0.004$)} \\
\midrule
$s$ & $m=500$ & $m=1000$ & $m=2000$ & $m=4000$ \\
\midrule
0.05 & 0.169$\pm$0.009 & 0.169$\pm$0.003 & 0.178$\pm$0.006 & 0.186$\pm$0.002 \\
0.10 & 0.200$\pm$0.001 & 0.199$\pm$0.006 & 0.193$\pm$0.010 & 0.189$\pm$0.008 \\
0.20 & 0.198$\pm$0.003 & 0.205$\pm$0.003 & 0.200$\pm$0.005 & 0.204$\pm$0.001 \\
\midrule
\multicolumn{5}{c}{$k=64$ (SimHash baseline: $0.177 \pm 0.004$)} \\
\midrule
$s$ & $m=500$ & $m=1000$ & $m=2000$ & $m=4000$ \\
\midrule
0.05 & 0.434$\pm$0.014 & 0.415$\pm$0.006 & 0.379$\pm$0.005 & 0.378$\pm$0.006 \\
0.10 & 0.446$\pm$0.004 & 0.456$\pm$0.003 & 0.451$\pm$0.009 & 0.442$\pm$0.014 \\
0.20 & 0.453$\pm$0.005 & 0.472$\pm$0.001 & 0.472$\pm$0.003 & 0.472$\pm$0.004 \\
\midrule
\multicolumn{5}{c}{$k=256$ (SimHash baseline: $0.469 \pm 0.003$)} \\
\midrule
$s$ & $m=500$ & $m=1000$ & $m=2000$ & $m=4000$ \\
\midrule
0.05 & 0.588$\pm$0.007 & 0.667$\pm$0.003 & 0.684$\pm$0.007 & 0.632$\pm$0.011 \\
0.10 & 0.607$\pm$0.004 & 0.683$\pm$0.002 & 0.707$\pm$0.006 & 0.689$\pm$0.007 \\
0.20 & 0.609$\pm$0.003 & 0.686$\pm$0.002 & 0.712$\pm$0.003 & 0.719$\pm$0.002 \\
\bottomrule
\end{tabular}
\end{table}

\begin{figure}[h]
\centering
\includegraphics[width=\textwidth]{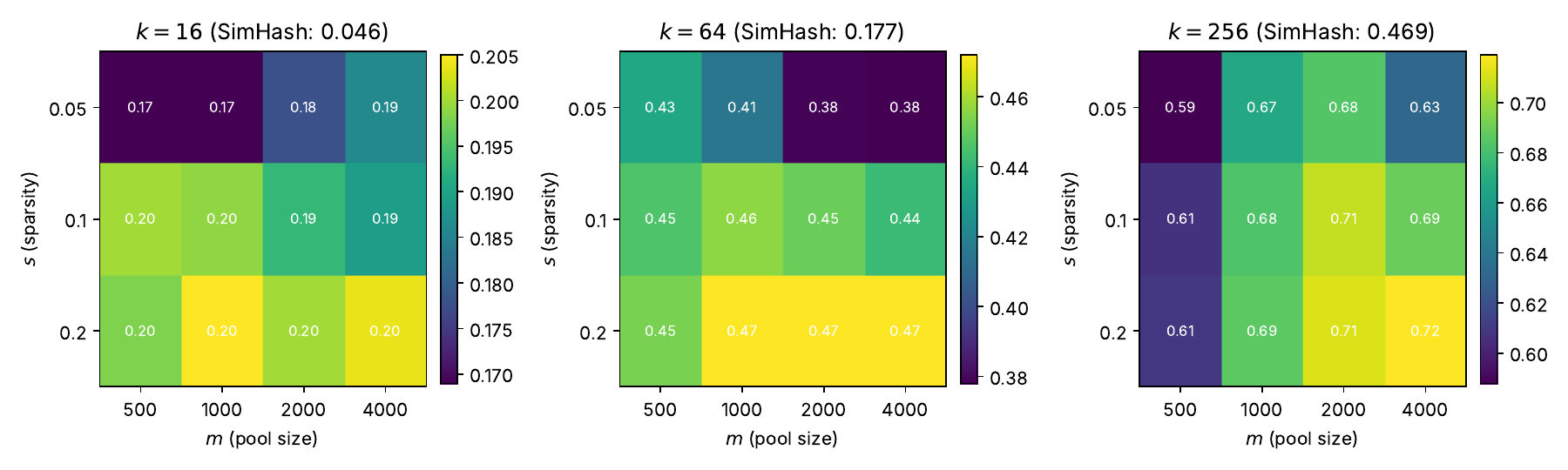}
\caption{FlyHash mAP@100 across the sparsity $\times$ pool-size grid, at
$k=16, 64, 256$. Each panel's SimHash baseline is given in its title;
every cell in every panel is above that baseline. The rightward
brightening within the $k=256$ panel visualizes the pool-size effect, and
the top-to-bottom brightening within each panel visualizes the sparsity
effect.}
\label{fig:heatmap}
\end{figure}

All 36 configurations exceed the SimHash baseline at their respective bit
budget --- no combination of sparsity and pool size reverses the result.
Two secondary patterns are visible: sparsity somewhat above the fly's own
biological ratio (0.20 vs.\ $\approx$0.10--0.12) performs marginally
better at larger bit budgets, and increasing pool size beyond roughly
$8\times$ the bit budget yields diminishing or slightly negative returns.
Neither pattern threatens the main result; both are minor tuning
observations. Figure~\ref{fig:heatmap} renders Table~\ref{tab:hyperparam}
as heatmaps, which makes both secondary patterns easier to see at a
glance than the raw numbers.

\subsection{Memory and query latency: quality gains do not imply resource gains}
\label{sec:memlatency}

The results above measure retrieval \emph{quality} only. We separately
measure memory footprint and query latency on GloVe (5{,}000 records),
comparing FlyHash and SimHash at matched $k$ against an exact (unhashed)
Euclidean nearest-neighbor baseline, using brute-force comparison for all
three (no approximate index on either side).

\begin{table}[h]
\centering
\caption{Packed memory and mean query latency (50 queries, brute-force
ranking against all 5{,}000 records). Exact float32 baseline: 400
bytes/record, 0.78 ms/query.}
\label{tab:memlatency}
\begin{tabular}{cccc}
\toprule
$k$ (bits) & SimHash bytes/record & FlyHash (dense) ms/query & SimHash ms/query \\
\midrule
16  & 2  & 14.05 & 0.19 \\
32  & 4  & 12.76 & 0.29 \\
64  & 8  & 12.93 & 0.45 \\
128 & 16 & 14.15 & 0.84 \\
256 & 32 & 13.17 & 2.17 \\
\bottomrule
\end{tabular}
\end{table}

FlyHash's code is not itself $k$ bits wide --- it is $m$ bits wide
(2{,}000 in our default configuration), of which exactly $k$ are active.
Our default dense representation stores and compares this full $m$-bit
array regardless of $k$, which makes FlyHash's packed size a
near-constant $\sim$250 bytes/record independent of $k$ --- larger than
SimHash's genuinely $k$-bit code at every budget tested except
$k \gtrsim 250$ --- and its brute-force query time a near-constant
$\sim$13 ms/query, 10--70$\times$ slower than SimHash at the same $k$.
This is a real structural property of the dense representation, not a
bug: mAP depends only on \emph{which} $k$ bits are active, so it is
entirely unaffected, but memory and speed under this representation
clearly are.

\subsection{A sparse-set comparison: partial recovery}
\label{sec:sparse}

Because every FlyHash code has exactly $k$ active bits, the Hamming
distance between two such codes reduces to $2k - 2|A \cap B|$, where $A$
and $B$ are their sets of active indices. Ranking by Hamming distance is
therefore identical to ranking by descending intersection size, which we
compute as a sparse-sparse matrix product over only the active indices,
rather than a dense operation over the full $m$-wide array.

\begin{table}[h]
\centering
\caption{Sparse vs.\ dense FlyHash comparison, and memory as an explicit
index list (uint16 per active index).}
\label{tab:sparse}
\small
\begin{tabular}{@{}cccccc@{}}
\toprule
\multirow{2}{*}{$k$ (bits)} & Sparse-idx & SimHash & Dense & Sparse & SimHash \\
 & bytes/rec & bytes/rec & ms/query & ms/query & ms/query \\
\midrule
16  & 32  & 2  & 15.77 & 0.71 & 0.40 \\
32  & 64  & 4  & 14.44 & 0.73 & 0.29 \\
64  & 128 & 8  & 14.54 & 1.59 & 0.45 \\
128 & 256 & 16 & 15.03 & 3.39 & 0.84 \\
256 & 512 & 32 & 13.90 & 8.55 & 1.79 \\
\bottomrule
\end{tabular}
\end{table}

The sparse comparison is 10--22$\times$ faster than the dense comparison
and, unlike the dense version, its latency correctly scales with $k$
rather than remaining flat. It does not, however, close the gap to
SimHash entirely: SimHash's small dense $k \times k$ matrix
multiplication remains 2--5$\times$ faster at every budget tested, and
explicit index-list storage, while now also correctly scaling with $k$,
remains $\sim$16$\times$ larger than SimHash's bit-packed code, since
storing an explicit position costs a full integer where bit-packing
costs a single bit. We report this as a genuine, partial improvement ---
not a solved problem --- and discuss the remaining gap in
Section~\ref{sec:limitations}.

\subsection{Agent memory demonstration}
\label{sec:agent-demo}

The end-to-end system described in Section~\ref{sec:agent-method}
functions correctly: in the representative interaction described there,
the agent retrieves a fact shared several turns earlier via
FlyHash-based Hamming ranking over MiniLM sentence embeddings, without
that fact appearing anywhere in the immediate conversational context.
This confirms the pipeline is functional for its intended use case,
independent of the benchmark numbers in
Sections~\ref{sec:main-result}--\ref{sec:sparse}.

As one measured data point for the full pipeline's latency (embedding
included, not just the isolated hash comparison of
Sections~\ref{sec:memlatency}--\ref{sec:sparse}), we timed a single
representative query against the 6,000-record corpus of
Section~\ref{sec:realistic-benchmark}, averaged over 30 repeats on
consumer hardware, using the sparse-set comparison method of
Section~\ref{sec:sparse} for the hash-ranking step (the same method
behind Table~\ref{tab:sparse}'s speed numbers, consistent with the rest
of the paper rather than an unrelated, unoptimized implementation):

\begin{table}[h]
\centering
\caption{End-to-end pipeline latency, one representative query (30 repeats).}
\label{tab:e2e-latency}
\small
\begin{tabular}{lrr}
\toprule
Stage & ms/query & \% of total \\
\midrule
Sentence embedding (MiniLM) & 14.1 & 68.8\% \\
FlyHash hash + sparse rank & 6.4 & 31.2\% \\
\textbf{Total end-to-end} & \textbf{20.5} & 100\% \\
\bottomrule
\end{tabular}
\end{table}

With the hash-ranking step measured consistently with
Table~\ref{tab:sparse}, sentence embedding --- not FlyHash's hash
comparison --- is the dominant cost of the full pipeline, by roughly
2:1. This is a useful finding in its own right: it means that, at this
corpus size, FlyHash's own contribution to end-to-end latency is modest
relative to the (hash-independent) embedding step, and that further
latency optimization in a real deployment would pay off more from the
embedding stage than the hash comparison. Embedding time is identical
regardless of which hash is used downstream, so this number has no
bearing on the FlyHash-vs-SimHash comparisons elsewhere in the paper ---
it exists only to put a real figure on the full pipeline's latency
rather than leaving it unmeasured. We still report it as a single data
point on one machine, not a benchmark: no seed averaging and no
variation in corpus size or hardware. A proper system-level benchmark
remains future work (Section~\ref{sec:limitations}).

\subsection{Practical systems comparison: FlyHash/SimHash vs. a graph-based ANN index}
\label{sec:hnsw-systems}

Sections~\ref{sec:main-result}--\ref{sec:sparse} hold the active-bit
budget $k$ fixed and compare FlyHash against SimHash --- both compact
hash codes compared by Hamming distance. This answers ``which hash is
better at a given bit budget,'' but not ``would I actually use either of
these instead of a standard production ANN index.'' We report that
second, separate question here using FAISS's \texttt{IndexHNSWFlat}
($M=32$, efConstruction$=200$) over full-precision GloVe vectors (5,000
records, 50 queries, mAP@100), explicitly as a systems comparison rather
than a matched-bit one: HNSW has no ``bits'' parameter comparable to
$k$, so its quality/speed/memory trade-off is governed by efSearch
instead, and the two tables below should be read within themselves, not
row-matched against each other.

\begin{table}[h]
\centering
\caption{FAISS HNSW, varying efSearch (exact L2 baseline: 0.909 ms/query, 400 bytes/record).}
\label{tab:hnsw}
\small
\begin{tabular}{rlll}
\toprule
efSearch & mAP@100 & ms/query & bytes/record \\
\midrule
16  & 0.7700 & 0.029 & 671.5 \\
64  & 0.9862 & 0.060 & 671.5 \\
128 & 0.9986 & 0.092 & 671.5 \\
\bottomrule
\end{tabular}
\end{table}

\begin{table}[h]
\centering
\caption{FlyHash (sparse-set) and SimHash, matched active-bit $k$, same dataset/queries.}
\label{tab:hnsw-hash}
\small
\begin{tabular}{rlllll}
\toprule
$k$ (bits) & FlyHash mAP & FlyHash ms/q & FlyHash B/rec & SimHash mAP & SimHash ms/q \\
\midrule
16  & 0.198 & 0.66  & 32  & 0.048 & 0.19 \\
32  & 0.312 & 1.04  & 64  & 0.095 & 0.22 \\
64  & 0.440 & 1.62  & 128 & 0.185 & 0.26 \\
128 & 0.582 & 4.46  & 256 & 0.325 & 0.50 \\
256 & 0.697 & 10.13 & 512 & 0.476 & 0.70 \\
\bottomrule
\end{tabular}
\end{table}

At this scale, HNSW dominates both hash methods on quality and latency
simultaneously --- mAP above 0.98 and sub-0.1ms queries, at only
$\sim1.6\times$ the raw exact-storage memory cost (671.5 vs.\ 400
bytes/record) --- while FlyHash and SimHash's only genuine advantage is
extreme compactness well below exact storage (as little as 2--32
bytes/record). The honest conclusion is that a practitioner with no
unusual memory constraint and a dataset at this scale should use HNSW,
not a hash code; FlyHash/SimHash's relevance is specifically to the
regime HNSW does not target --- severe, explicit storage budgets (e.g.,
many independent memory stores on resource-constrained agents) --- and
we revise our framing in Section~\ref{sec:discussion} accordingly rather
than letting Sections~\ref{sec:main-result}--\ref{sec:sparse} imply a
general retrieval recommendation.

\subsection{A realistic, labeled agent-memory benchmark}
\label{sec:realistic-benchmark}

Every result above uses nearest-neighbor-by-embedding as a stand-in for
``relevance'' --- two items are treated as relevant to each other if an
unhashed embedding considers them close, which is circular with respect
to the embedding itself and does not test anything about whether hashing
preserves \emph{true} query-to-memory relevance. We address this with a
separate benchmark using constructed relevance labels that do not derive
from any embedding distance.

Established agent-memory benchmarks such as LoCoMo (Maharana et al., 2024)
and LongMemEval (Wu et al., 2024) evaluate end-to-end conversational QA
accuracy over multi-session dialogue, which conflates retrieval quality
with the downstream LLM's reasoning over retrieved context (Wu et al.,
2024, explicitly separate indexing, retrieval, and reading for this
reason). Our evaluation instead requires ground-truth relevance labels for
the retrieval step in isolation --- the quantity FlyHash and SimHash
actually compete on --- which these QA-style benchmarks do not provide
directly. We therefore construct a dataset with explicit, non-circular
relevance labels anchored to named personas below, in the same spirit as
recent calls for benchmark designs that disentangle retrieval correctness
from answer correctness (Wu et al., 2024; Hao et al., 2026).

\textbf{Construction.} We generate 150 synthetic named personas, each
with 10 independent facts (favorite color, allergy, job title, hometown,
and six others), each fact expressed via 3--4 paraphrase templates that
name the persona directly (e.g., ``Jordan's favorite color is teal.'' /
``If you ask Jordan, teal is their favorite color.''), giving 6,000
memory sentences total. Each query is a differently-worded question that
also names a target persona (e.g., ``Which color does Jordan like
best?''), and its ground-truth relevant set is exactly that persona's own
paraphrases for that fact (3--4 sentences out of 6,000) --- a label
fixed by construction, not by embedding similarity. Naming every fact
and question to a specific persona is what makes the corpus both
genuinely confusable (thousands of near-identical same-template
statements about \emph{other} people crowd every query) and answerable
from text content alone (the one token that correctly disambiguates the
right answer --- the name --- is actually present in the query), which
an earlier, unnamed version of this design was not: without a persona
name tying question to answer, no text-similarity method, hashed or not,
can identify \emph{whose} stated value is correct among many textually
identical same-template statements, and we do not report numbers from
that earlier, ill-posed version.

\textbf{Embedding.} We embed both memories and queries with TF-IDF
followed by truncated SVD (256 components), fit on the memory corpus ---
an offline, network-free stand-in for a real sentence-transformer
embedding, consistent with this project's practice elsewhere
(Section~\ref{sec:limitations}).

\begin{table}[h]
\centering
\caption{FlyHash vs.\ SimHash, standard mean Average Precision against constructed relevance labels (150 queries, 6,000 memories, 10 seeds).}
\label{tab:realistic-benchmark}
\small
\begin{tabular}{rlll}
\toprule
$k$ (bits) & FlyHash mAP (labeled) & SimHash mAP (labeled) & Robust \\
\midrule
16  & \textbf{0.195 $\pm$ 0.014} & 0.038 $\pm$ 0.010 & Yes \\
32  & \textbf{0.271 $\pm$ 0.017} & 0.105 $\pm$ 0.011 & Yes \\
64  & \textbf{0.324 $\pm$ 0.012} & 0.212 $\pm$ 0.016 & Yes \\
128 & \textbf{0.365 $\pm$ 0.009} & 0.298 $\pm$ 0.011 & Yes \\
256 & \textbf{0.392 $\pm$ 0.009} & 0.345 $\pm$ 0.011 & Yes \\
\bottomrule
\end{tabular}
\end{table}

FlyHash beats SimHash robustly at every budget on this genuinely-labeled
task, with absolute mAP values well above chance (random ranking over
6,000 items with $\sim$4 relevant would score far below 0.1) but,
expectedly, lower than Table~\ref{tab:main-results}'s
nearest-neighbor-pseudo-label mAP --- this is a harder, more realistic
task by construction, and the two numbers are not meant to be compared
directly. This result is, to our knowledge, the first time FlyHash's
advantage over SimHash has been shown against actual, non-circular
relevance labels for an agent-memory-shaped task rather than an
embedding-similarity proxy.

\section{Discussion}
\label{sec:discussion}

\textbf{Why is the advantage largest at low bit budgets?} FlyHash's
sparse expansion into a much higher-dimensional space before selecting
winners appears to preserve more of the input's neighborhood structure
per output bit than SimHash's direct dense projection, particularly when
very few bits are available to encode that structure. As the bit budget
grows, both methods have enough capacity to represent more of the
relevant structure, and the relative gap narrows --- though FlyHash's
absolute advantage never disappears across any budget we tested
(Table~\ref{tab:main-results}). This pattern is practically significant:
real deployments of compact hashing are motivated precisely by wanting
\emph{few} bits per record (that is the entire point of hashing over
storing full embeddings), so the regime where FlyHash's advantage is
largest is also the regime a real system would actually operate in.

\textbf{A quality advantage is not automatically a resource advantage.}
Section~\ref{sec:memlatency} is, to our knowledge, an underappreciated
point in this line of work: a hash function can preserve neighbor
structure strictly better than a baseline while, in a naive
implementation, being simultaneously worse on memory and speed ---
because ``better structure per bit'' and ``cheap to store and compare''
are different properties that do not automatically follow from each
other. Section~\ref{sec:sparse} shows that a principled fix targeting the
comparison's sparsity recovers much, but not all, of one axis (speed)
while leaving the other (memory) only partially improved. We think this
distinction --- quality-per-bit versus deployed resource cost --- is
worth making explicit in any evaluation of a hashing scheme derived from
a sparse-coding argument, not only in this paper.

\textbf{Practical implication for agent memory.} An agent adopting
FlyHash today obtains a genuine, robustly validated retrieval-quality
improvement per bit over SimHash, particularly in the low-bit-budget
regime most relevant to a resource-constrained memory system, and
(Section~\ref{sec:realistic-benchmark}) that advantage survives contact
with genuine, non-circular relevance labels, not just an
embedding-similarity proxy. It should not, on the basis of this paper,
expect a memory or latency improvement over a well-implemented SimHash
baseline without further engineering (Section~\ref{sec:limitations}),
nor should it expect either hash method to beat a standard production
ANN index such as HNSW at ordinary scale (Section~\ref{sec:hnsw-systems})
--- HNSW wins on quality and latency simultaneously there, at a modest
memory premium over exact storage. Which of these properties matters
most depends on the deployment: a system already committed to compact
hash codes, bottlenecked on retrieval accuracy at a fixed, small bit
budget, benefits from FlyHash as a drop-in SimHash replacement; a system
with no unusual storage constraint should likely use HNSW instead of
either hash; and a system bottlenecked specifically on raw per-record
storage at a scale where even HNSW's overhead is unaffordable is the
narrow regime where FlyHash/SimHash-style hashing is the right tool at
all.

\section{Limitations and Future Work}
\label{sec:limitations}

\begin{itemize}[leftmargin=*]
\item \textbf{Memory and speed do not fully match the quality advantage,
even after a targeted attempt to close the gap.}
Section~\ref{sec:sparse}'s sparse-set comparison recovers most of the
speed disadvantage but leaves a meaningful memory disadvantage; a
genuinely sublinear index (e.g., banded or multi-probe LSH, or an
inverted index verified sublinear at scale --- an earlier inverted-index
attempt in our own implementation was found not to be sublinear at the
scales we tested) remains unbuilt. A more compact encoding of the
active-index list (delta-coding of sorted indices, variable-length or
entropy coding, or reducing pool size at some cost to quality) is a
concrete, identified next step we did not pursue here.
\item \textbf{Our HNSW comparison (Section~\ref{sec:hnsw-systems}) is a
single-scale, single-configuration snapshot.} We report one dataset
(GloVe, 5,000 records) and one HNSW configuration ($M=32$,
efConstruction$=200$, three efSearch values); HNSW's relative advantage
over hashing shrinks as per-record storage budgets tighten and may
shrink or reverse at very different corpus scales, which we have not
swept. A combined approach (e.g., FlyHash codes as a cheap pre-filter
ahead of an HNSW-indexed re-ranking stage, in the spirit of how FAISS
combines quantization with HNSW) is a natural direction we have not
explored.
\item \textbf{SIFT1M, part of the original paper's evaluation, is not
included.} We judged that four independent datasets across two
modalities already provide strong evidence the result is not
dataset-specific, and that a fifth, necessarily subsampled dataset
(SIFT1M's full 1M-vector scale is impractical to evaluate exhaustively at
our compute budget) would add breadth rather than materially new
evidence. We report this as a deliberate scope decision, not an
oversight.
\item \textbf{We do not implement DenseFly's densification or BioHash's
learned projection}, both of which are reported improvements over the
plain, random-projection FlyHash evaluated here
(Section~\ref{sec:related}). Combining either with the agent-memory
application is natural future work.
\item \textbf{We compare two hashing paradigms (sparse random-projection
vs.\ dense random-hyperplane) but do not benchmark against vector
quantization methods} (e.g., Product Quantization, RaBitQ), which the
literature indicates generally outperform hashing in accuracy-per-bit
(Jégou et al., 2011; Gao \& Long, 2024; Section~\ref{sec:related}); a
comparison against quantization is natural future work alongside the
HNSW baseline already included.
\item \textbf{The real-text (MiniLM) evaluation uses a small,
100-sentence corpus}, chosen to avoid duplicate-embedding artifacts (see
Appendix~\ref{sec:appendixA}) rather than to represent realistic
agent-memory scale. A larger, more diverse real-text corpus would
strengthen this specific result.
\item \textbf{We report only a single data point for the full
end-to-end agent memory system's latency} (Section~\ref{sec:agent-demo}:
20.5 ms/query on one machine, one query, 30 repeats), not a proper
benchmark --- no seed averaging and no sweep over corpus size or
hardware. A dedicated, multi-seed system-level benchmark is future work.
\item \textbf{The realistic, labeled agent-memory benchmark
(Section~\ref{sec:realistic-benchmark}) uses templated synthetic text,
not naturally-occurring language, and a TF-IDF+SVD embedding as a
stand-in for a real sentence-transformer.} Real sentence diversity is
considerably messier than any fixed template set, and real deployments
would more plausibly scope memory retrieval by session or user ID than
rely on a name mentioned in free text. A rerun with real human-written
memory logs and a real sentence-transformer embedding (on a machine with
internet access, the same gap already noted for MiniLM/MNIST elsewhere
in this paper) would strengthen this specific result.
\end{itemize}

\section{Conclusion}

Our three research questions have three different shapes of answer, and
we state the claim no wider than that. FlyHash beats SimHash at matched
active-bit counts, robustly, on every dataset and seed we tested,
including under genuine agent-memory relevance labels rather than an
embedding-similarity proxy (RQ1, RQ3) --- this is a real, well-supported
result, and one previously undocumented comparison pitfall ($k$-vs-$m$
conflation) had to be corrected to see it. But the practical deployment
question (RQ2) does not resolve as cleanly: FlyHash's quality advantage
over SimHash does not translate into a memory or latency advantage under
a naive comparison, a sparse-set comparison closes most but not all of
that gap, and a standard production index (HNSW) outperforms both hash
methods on every axis at ordinary scale, leaving genuinely extreme
storage constraints as hashing's specific niche rather than a general
deployment recommendation. We think this --- a validated hash-quality
result, a corrected methodological pitfall, and a resource picture that
narrows rather than flatters the headline claim --- is a more useful
contribution to reproducible progress here than a report that stopped at
the first, more flattering result.

\appendix

\section{Reproducibility Investigation}
\label{sec:appendixA}

While reproducing the original paper's reported result on GloVe word
vectors, we initially found FlyHash losing to SimHash on every dataset we
tried --- synthetic clustered embeddings, TF-IDF text vectors, real
sentence embeddings, and real GloVe vectors --- directly contradicting
the published finding on GloVe. We treated this discrepancy as a signal
that our comparison, not the algorithm, was most likely at fault, and
pursued it systematically rather than accepting or explaining away the
contradiction.

\textbf{Hypotheses ruled out.} We first tested whether L2-normalizing
inputs before hashing affected the result; it did not, meaningfully. We
then tested whether using the algorithm's exact original weight
definition (true binary $\{0,1\}$ projection weights, rather than the
signed $\{-1,+1\}$ weights we had implemented) affected the result; it
did not, and we quantify this claim across all five main bit budgets
below rather than resting it on the single seed we checked at the time.
We then tested whether our evaluation metric --- recall@$k$ by overlap,
rather than mAP@$R$ --- was responsible; switching to mAP@$R$ produced
the same qualitative picture. Each ruled-out hypothesis narrowed the
search space without resolving the discrepancy.

\textbf{Quantifying the weight-mode claim.} The ``it did not,
meaningfully'' judgment above was originally made on a single seed.
Re-run across all 10 seeds and all five main bit budgets (GloVe), with
SimHash included as a reference:

\begin{table}[h]
\centering
\caption{Signed vs.\ binary FlyHash projection weights, GloVe, 10 seeds.}
\label{tab:weight-mode}
\small
\begin{tabular}{rlllll}
\toprule
$k$ (bits) & Signed $\{-1,+1\}$ mAP & Binary $\{0,1\}$ mAP & $|$diff$|$ & SimHash mAP (ref.) \\
\midrule
16  & 0.196 $\pm$ 0.007 & 0.191 $\pm$ 0.005 & 0.0042 & 0.047 $\pm$ 0.006 \\
32  & 0.314 $\pm$ 0.009 & 0.298 $\pm$ 0.007 & 0.0156 & 0.096 $\pm$ 0.009 \\
64  & 0.449 $\pm$ 0.008 & 0.421 $\pm$ 0.005 & 0.0277 & 0.185 $\pm$ 0.008 \\
128 & 0.585 $\pm$ 0.006 & 0.540 $\pm$ 0.005 & 0.0449 & 0.316 $\pm$ 0.006 \\
256 & 0.705 $\pm$ 0.006 & 0.647 $\pm$ 0.005 & 0.0577 & 0.469 $\pm$ 0.007 \\
\bottomrule
\end{tabular}
\end{table}

Both weight modes beat SimHash by a wide margin at every budget tested.
The signed-vs-binary difference is not zero and grows with $k$ (0.004 at
$k=16$ to 0.058 at $k=256$), so the honest claim is not ``no difference''
but ``small relative to the FlyHash-vs-SimHash effect at every budget we
tested, with a modest upward trend at higher $k$'' --- the main results
in Section~\ref{sec:main-result} use the signed variant throughout, and
this table bounds how much of that result could plausibly be an
artifact of that choice.

\textbf{The actual cause.} We resolved the discrepancy by comparing our
implementation's parameterization directly against a third-party
reference implementation of the original algorithm
(\texttt{github.com/dataplayer12/Fly-LSH}). This revealed that our
comparison held FlyHash's \emph{total} hash dimensionality equal to
SimHash's bit count while separately fixing FlyHash's winner-take-all
\emph{percentage} --- silently making FlyHash's \emph{effective} number
of active bits, at any given nominal ``bits'' value, a small fraction of
what SimHash was being given as its full, genuinely active bit count.
This is precisely the $k$-versus-$m$ conflation described in
Section~\ref{sec:protocol}. Correcting the protocol to hold $k$ (not $m$)
equal between the two methods, with $m$ fixed independently at a large
value, immediately reproduced the original paper's reported pattern:
FlyHash winning decisively on GloVe at every bit budget, and --- as it
turned out --- on every other dataset we subsequently tested as well
(Section~\ref{sec:main-result}).

\textbf{A hypothesis retracted.} Prior to finding this cause, we had
observed that FlyHash appeared to outperform SimHash specifically on
data with correlated, non-negative features (e.g., TF-IDF text vectors)
but not on isotropic synthetic data, and had proposed that FlyHash
specifically benefits from such data properties. Once the actual
comparison bug was corrected, this pattern disappeared: FlyHash
outperformed SimHash on the corrected synthetic isotropic data just as
decisively as on text data. We state plainly that our earlier hypothesis
was incorrect --- it was constructed to explain an artifact of a flawed
comparison, not a genuine property of either the data or the algorithm
--- and we retract it here rather than omit it from this account.

We document this investigation in this level of detail because the
underlying pitfall --- conflating a hash's \emph{effective} bit count
with its \emph{nominal} parameter dimensionality --- seems easy to make
and is, to our knowledge, not explicitly warned against elsewhere in this
literature. We hope stating it precisely, alongside the specific
hypotheses that seemed plausible but were not the actual cause, helps
others attempting a similar comparison avoid the same error.

\end{document}